\documentclass{bmvc2k}
\usepackage[table]{xcolor}
\usepackage{multirow, tabularx}
\usepackage{booktabs}
\usepackage{adjustbox}
\usepackage{kotex}
\usepackage{array}

\title{Superpixel-based semantic segmentation trained by statistical process control}

\addauthor{Hyojin Park}{wolfrun@snu.ac.kr}{1}
\addauthor{Jisoo Jeong}{soo3553@snu.ac.kr}{1}
\addauthor{Youngjoon Yoo}{i0you200@snu.ac.kr}{1}
\addauthor{Nojun Kwak}{nojunk@snu.ac.kr}{1}

\addinstitution{
 Department of Transdisciplinary Studies, \\
 Seoul National University, \\
 Republic of Korea.
}

\runninghead{H. Park, J. Jeong, Y. Yoo and N. Kwak}{HP-SPS }

\def\etal{\emph{et al}\bmvaOneDot}

\begin{document}
\maketitle

\begin{abstract}
Semantic segmentation, like other fields of computer vision, has seen a remarkable performance advance by the use of deep convolution neural networks.
However, considering that neighboring pixels are heavily dependent on each other, both learning and testing of these methods have a lot of redundant operations. 
To resolve this problem, the proposed network is trained and tested with only 0.37\% of total pixels by superpixel-based sampling and largely reduced the complexity of upsampling calculation. 
%
The hypercolumn feature maps are constructed by pyramid module in combination with the convolution layers of the base network.
Since the proposed method uses a very small number of sampled pixels, the end-to-end learning of the entire network is difficult with a common learning rate for all the layers. 
In order to resolve this problem, the learning rate after sampling is controlled by \textit{statistical process control} (SPC) of gradients in each layer. The proposed method performs better than or equal to the conventional methods that use much more samples on \textit{Pascal Context, SUN-RGBD} dataset.
\end{abstract}


\section{Introduction}
\label{sec:intro}

\begin{figure}[t]
\small
\centering
\begin{adjustbox}{width=\textwidth}
\begin{tabular}{cc}

\bmvaHangBox{\includegraphics[width=6cm, height=3cm]{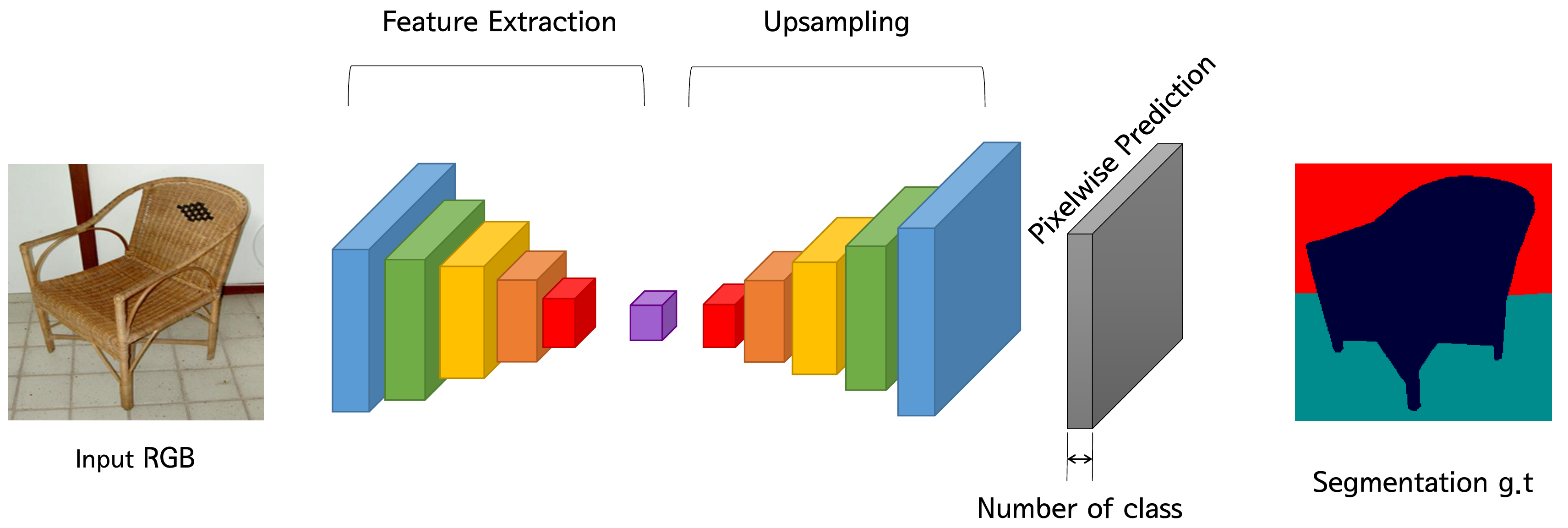}}&
\bmvaHangBox{{\includegraphics[width=6cm, height=3cm]{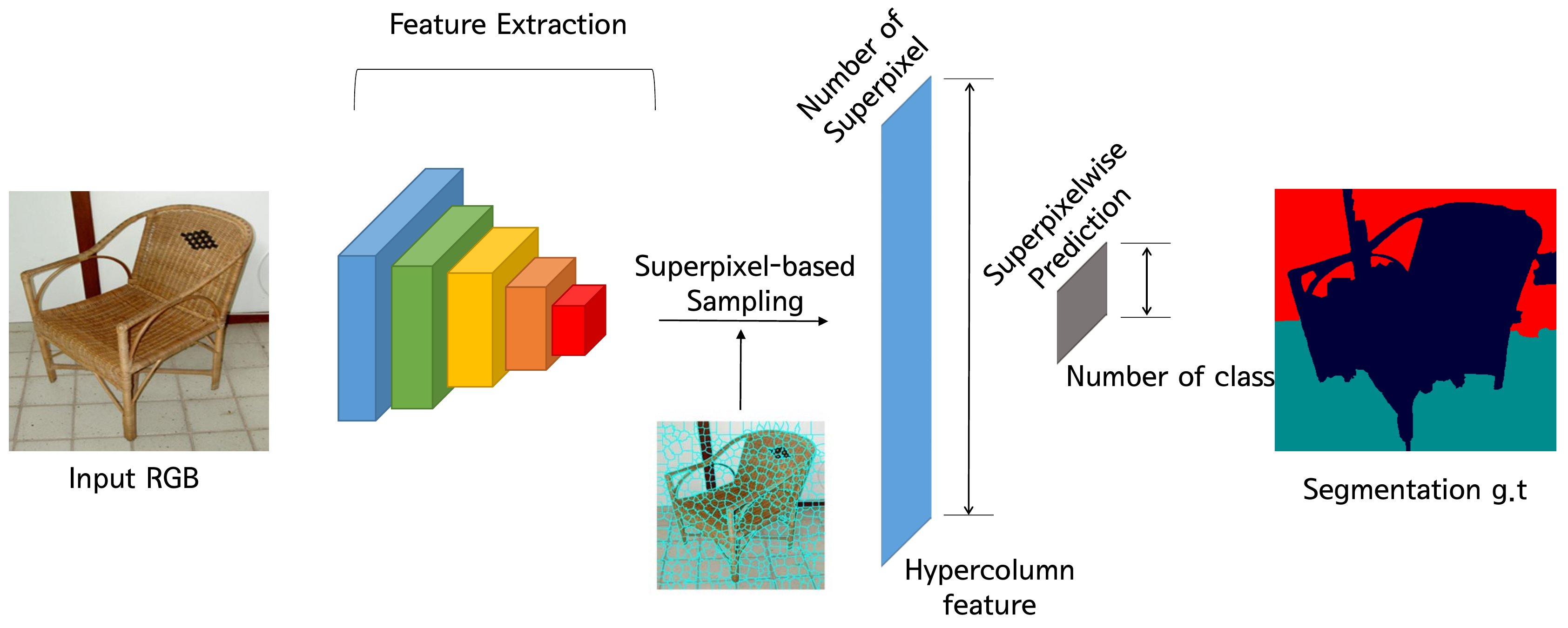}}}\\
(a)&(b)

\end{tabular}
\end{adjustbox}
\caption{ (a) Conventional CNN for semantic segmentation. Usually they need upsampling layer and do pixel-wise classification (b) Our proposed method. We do not need to use upsampling method and reduce the operations significantly  by superpixel based sampling } 
\label{fig:fig1}
\end{figure}

The purpose of semantic segmentation is to segment a given image and identify the semantic information of each segment. 
Like many other computer vision applications, 
architectures based on convolutional neural networks (CNN) have been introduced and applied to improve performance of the semantic segmentation \cite{shelhamer2017fully, shuai2016improving, farabet2013learning}.  Especially, since the introduction of the \textit{fully convolutional network} (FCN) based architecture proposed in \cite{shelhamer2017fully}, which showed promising performance on semantic segmentation, many studies follow this methodology \cite{YuKoltun2016, shuai2016improving, badrinarayanan2015segnet, noh2015learning}. A general CNN-based semantic segmentation is largely divided into three parts as shown in Figure \ref{fig:fig1}(a). First, feature extraction is performed as in \cite{Simonyan14c}.
Second, we upsample the reduced feature map to the original size and finally calculate probability values for each semantic class using a 1-by-1 convolution for  each pixel.
This is also referred to as dense classification or pixel-wise classification. 
\setlength{\textfloatsep}{5pt}

\begin{figure}[t]

\centering
\includegraphics[width=0.8\linewidth]{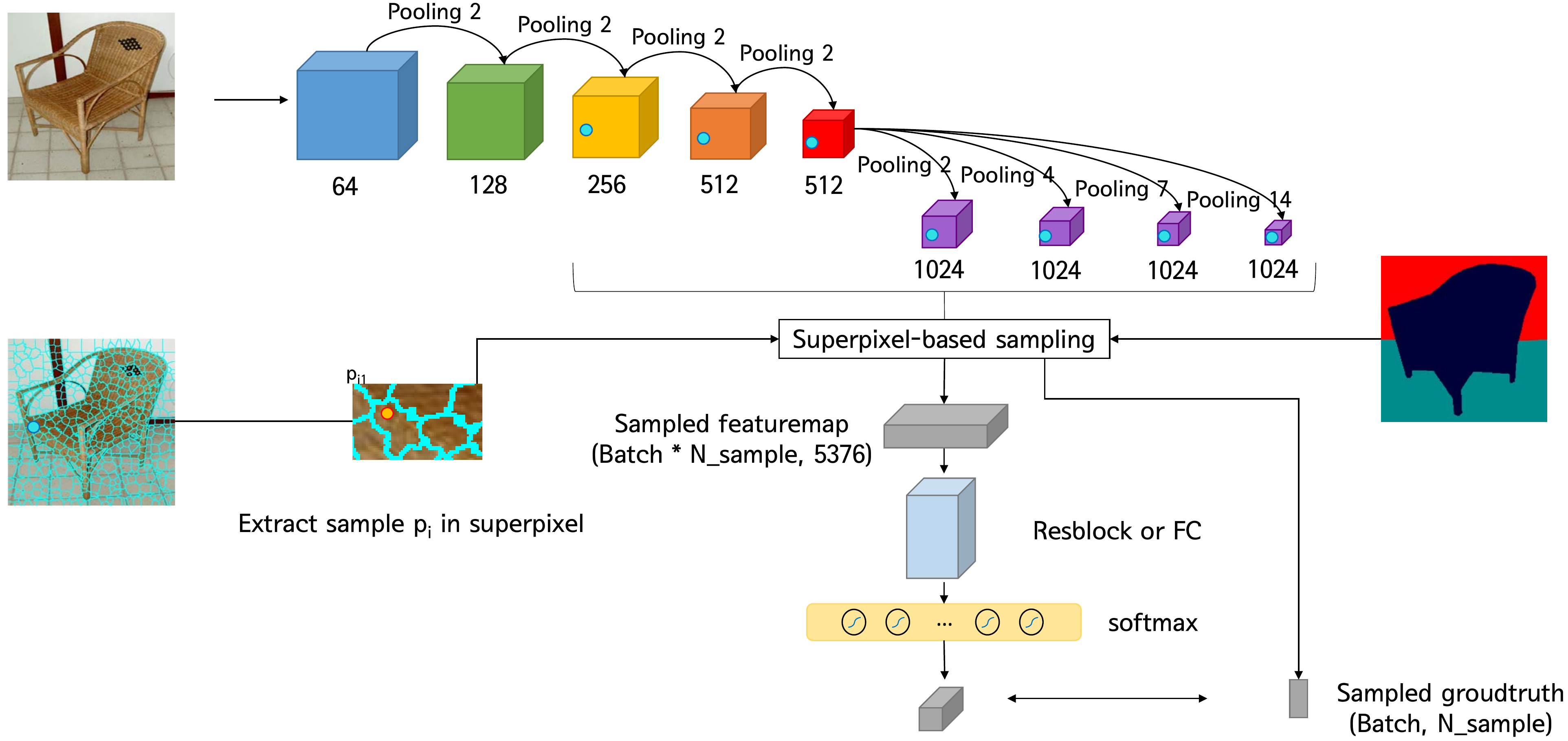}


\caption{ Overall structure of the proposed method (HP-SPS) }
\label{fig:overview}
\end{figure}
 \setlength{\textfloatsep}{5pt}
Semantic segmentation has a couple of problems owing to the pixel-wise classification. 
First, pooling is used multiple times for a wider receptive field to get richer information  as in \cite{bansal2017pixelnet}. 
Thus, the feature map becomes much smaller than the original size. 
In order to solve this problem, researchers devised schemes which used less pooling layers but with a wider receptive field, and made use of CNN-based upsampling methods.
 Second, neighboring pixels are likely to have close relationship and share similar information. Therefore, they are likely to belong to the same semantic class. 
However, the class of a pixel is calculated independently without this relation. 
If we group the pixels that are likely to belong to the same class, we can reduce the inefficient computation  and time complexity. 
Also, because the neighboring pixels share information, the basic assumption of the stochastic gradient descent (SGD), that the data have independent identical distribution (IID), is violated and the learning becomes inefficient  \cite{bottou2010large,hyvarinen2009natural,lecun2012efficient}. 

However, none of the above studies has solved the problems mentioned above completely. 
The use of upsampling 
requires additional computation, 
even though less pooling is performed. In addition, there is still a problem of pixel-wise classification. 
This paper proposes a novel semantic segmentation algorithm based on pyramid module in combination with superpixel-based sampling to solve the problems mentioned above. More specifically,
1) we use the pyramid module to broaden the receptive field and enable feature extraction with enhanced scale-invariant properties. 
2) We do not use common upsampling method \cite{hariharan2015hypercolumns} and changed the method more efficiently by applying superpixel-based sampling method 
that reduces the amount of computation in the training and testing, by using only 0.37\% of the total pixels through sampling. 
3) To solve the learning speed drop problem cased by the sampling, the learning rate is further tuned to acquire stable gradients, by using statistical process control \cite{shewhart1931economic}.

The organization of the paper is given as follows. In Section \ref{sec:related}, the related works are briefly overviewed. Section \ref{sec:method} presents the proposed method of semantic segmentation using superpixel-based sampling with pyramid module and hypercolumn feature extraction. We also propose new gradient control method to compensate for the learning problem in sampled networks. Section \ref{sec:exp} shows the experimental results using Pascal context with subtraction experiment and additional experiment using SUN-RGBD finally Section \ref{sec:conclusion} concludes the paper.

\section{Related Works}
\label{sec:related}
The performance of semantic segmentation has improved a lot since the introduction of CNN-based  methodologies \cite{shelhamer2017fully, farabet2013learning, shuai2016improving}. 
Since the introduction of the FCN \cite{shelhamer2017fully}, the scale invariant features could be obtained in a fast and easy way without the use of the image pyramid and semantic segmentation has been conducted using the popular networks such as VGGNet \cite{Simonyan14c}. After then, in many methods\cite{noh2015learning, chen2016deeplab, badrinarayanan2015segnet2}, pooling was used to obtain features with wider receptive fields \cite{chen2016deeplab}. 
However, successive application of pooling makes the resolution of a feature smaller and it becomes quite difficult to recover the original resolution of an image. 
To resolve this problem and to obtain features with wider receptive fields using the reduced number of pooling, some new types of filters such as dilated convolution \cite{YuKoltun2016} and atrous convolution \cite{chen2016deeplab} have been introduced. 
 
Nonetheless, recovering the resolution of feature map and preserving fine-detail information 
a major obstacle for the FCN based segmentation. Therefore, researchers have proposed the methods concatenating the features in intermediate layers as well as the final layer to obtain high quality features. The studies in \cite{badrinarayanan2015segnet, badrinarayanan2015segnet2, noh2015learning, Zagoruyko2016Multipath} belong to this line of research.
The hypercolumn \cite{hariharan2015hypercolumns} of a pixel is defined as a stacked vector of all features in feature maps in every different layers corresponding to the pixel. The methods of \textit{shift and stitch} \cite{sermanet2013overfeat,shelhamer2017fully}, \textit{deconvoution} \cite{noh2015learning}, and \textit{unpooling} \cite{badrinarayanan2015segnet, badrinarayanan2015segnet2, noh2015learning} gradually recover the missing information by adding extra shallow layer features of the same target resolution \cite{Zagoruyko2016Multipath}.

There are also studies different from this framework of \textit{extract-and-expand} as in \cite{shelhamer2017fully}. 
Lin \etal \cite{lin2016efficient} combine the structure of conditional random field (CRF)  and CNN, where the CNN computes the potential function value through joint learning. Then, a mean-field approximation is applied to perform semantic segmentation. In \cite{bansal2017pixelnet}, they randomly sampled the feature map to cut off dependency among the pixels and delivered the gradient only to the sampled ones for statistical efficiency. However, at the time of inference, they must create hypercolumn feature maps that are in the same size as the original image. Also, at the time of training, they use as much as 10 times the number of samples than ours. Therefore very redundant operation is still performed in \cite{bansal2017pixelnet}. In \cite{mostajabi2015feedforward}, multi-layer features were created by average pooling in each superpixel for semantic segmentation. This study aims at rich feature representation for various resolutions without using a complex additional model. At the same time, it also enhances efficiency in testing by using superpixel. 
Although this work has some commonality with ours in that it uses superpixels in training and testing, the difference is that unlike theirs, 1) we used sampling to reduce the computational complexity and 2) hypercolumns of pyramid modules for better representation power. 

\section{Proposed Method}
\label{sec:method}

Figure \ref{fig:overview} shows the overall framework of the proposed semantic segmentation method with superpixel-based sampling. 
First, the CNN features are extracted for a given image.
Second, the image is divided into small regions by using a superpixel technique. 
Third, each region in a superpixel is represented by one or two random pixels in the region.
Fourth, for each sampled pixel, hypercolumn features are generated by concatenating all the feature maps for each layer passing the pixel.
Then, using the extracted hypercolumn features, segmentation is conducted for each sampled pixel in a superpixel by using the segmentation network composed of Resblock \cite{he2016deep} or FCN. The proposed method is named as HP-SPS, an abbreviation for \textit{Hypercolums of Pyramid module with SuperPixel-based Sampling}. The detailed network design can be found in Table \ref{table1}. More detailed explanation of each step is given below.

\subsection{Feature Extraction through Hypercolumns of Pyramid Module}
\label{sec:prop1}

\begin{figure}[tbp]
\centering
\includegraphics[width=0.8\linewidth]{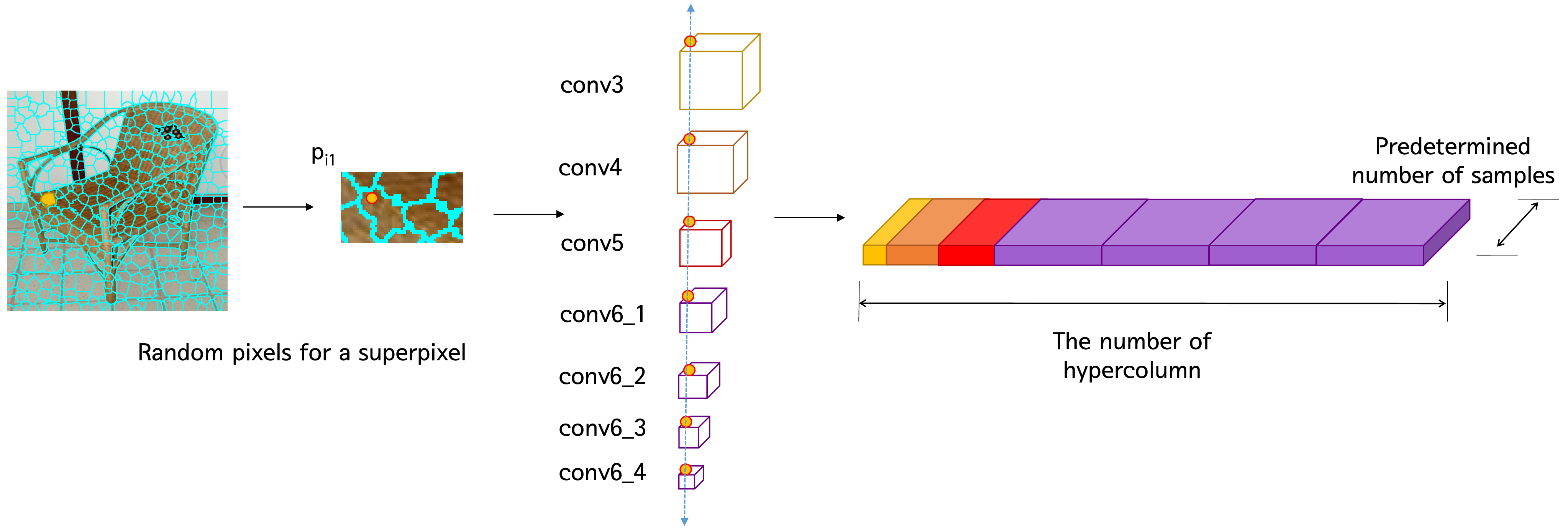}
\caption{Hypercolumns of random pixels}
\label{fig3}
\end{figure}

For semantic segmentation, we first map the input images into multi-layer CNN features having large receptive fields in order to make the features robust to scale and translation variations as in \cite{bansal2017pixelnet}.
In the proposed feature network, we use the VGGnet \cite{Simonyan14c} until conv5 stage.
Then, the receptive field is expanded by four parallel pooling layers with pool sizes of 2, 4, 7, and 14, respectively.
After pooling, $3\times 3$ convolution with 1024 dimension 
is performed on each output of the four pooling. We found that the segmentation performance is better when the kernel size is equivalent to the stride size, so that the convolution intervals do not overlap. Also kernel size of 3 was better than that of 1 because it considers the information of surrounding features. 
We set the output feature dimensions the same (1,024) because if the feature dimensions of the layers are much different from each other, it cause to use only specific scale information \cite{pinheiro2016learning}.
After mapping the image into the feature extraction network, we concatenate the feature maps for each layers using hypercolumn method \cite{hariharan2015hypercolumns}. To apply the method, we track the pooling locations of the target pixel through conv3, conv4, conv5 and all the four conv6 in the proposed feature extraction network, then concatenate all the corresponding feature maps of different layers as shown in Figure \ref{fig3}. 
We note that the normalization step is required to balance the scale between the layers. $l_2$ normalization is adopted in our method as in \cite{liu15parsenet}. 

\subsection{Super-pixel Based Sampling for Learning and Inference}
\label{sec:prop2}
In our method, superpixel is adopted as a basic block for semantic segmentation to resolve the problem of redundancy in learning and prediction. 
We weakly segment the image using simple Linear Iterative Clustering (SLIC) and randomly sample the representative pixel $p_i$ for each superpixel $s_i,i=1,\ldots,N$. 
Since the number of SLIC superpixels differs from image to image, we randomly select one or two pixels from the randomly chosen superpixel $s_i$ such that the number of the total selected pixels be same for all images. Then, we extract the hypercolumn feature $f_i$ for the selected pixel $p_i$ representing the superpixel $s_i$ and also record the segmentation label $l_i$ at same position as in Section \ref{sec:prop1}. By selecting both the superpixel and the pixels in it randomly, we try to meet the IID assumption of SGD.
 
Using the extracted hypercolumn feature $f_i$, we train the segmentation network in Figure \ref{fig:overview} with label $l_i$ for the superpixel $s_i$. For the segmentation network, Resblock \cite{he2016deep} and FCN \cite{shelhamer2017fully} are adopted. At test time, we assign the same class to all the pixels inside of superpixel for dense prediction. Unlike other studies, thanks to superpixel-based sampling, we do not need to recover feature map to the original resolution. Also because most of other works perform pixel-wise independent estimation with 1-by-1 convolution, they incur much computational complexity. Because neighbor pixels have high probability of sharing the same semantic information, our superpixel based sampling method reduces this complexity a lot. 
 
However, in our method, the training is not straightforward because the number of the training sample is drastically decreased (only 0.374\% of the sample compared to the pixel-wise case). This is severe considering the large dimension of the input $f_i$.
To train the network with a relatively small number of samples, learning rate of each layer $\gamma$ of the stochastic gradient should be increased so that the network parameter can be changed enough to reflect the effect of the sample $p_i$.
When increasing the learning rate, correspondingly, we suffer from noisy gradient problem and it is critical especially for the case when just a few input samples are provided. Therefore, we propose statistical process control (SPC) method to control the noisy gradient problem and successfully train the network parameters using a restricted number of input samples. We analyze the gradients from the experiments with low and high learning rates for all the layers (\textit{all low $\gamma$} and \textit{all high $\gamma$}, respectively) and suggest a hybrid learning rates method where in some layers $\gamma$ is set to have low values, while in other layers it is set to have high learning rate (\textit{hybrid $\gamma$}). 
Note that the control of $\gamma$ by SPC is applied only to the layers after the sampling.

\subsection{Statistical Process Control for Tuning Learning Rates}
\label{sec:prop3} 
\begin{figure}[tb]
\centering 
\begin{adjustbox}{width=0.8\textwidth}
\includegraphics{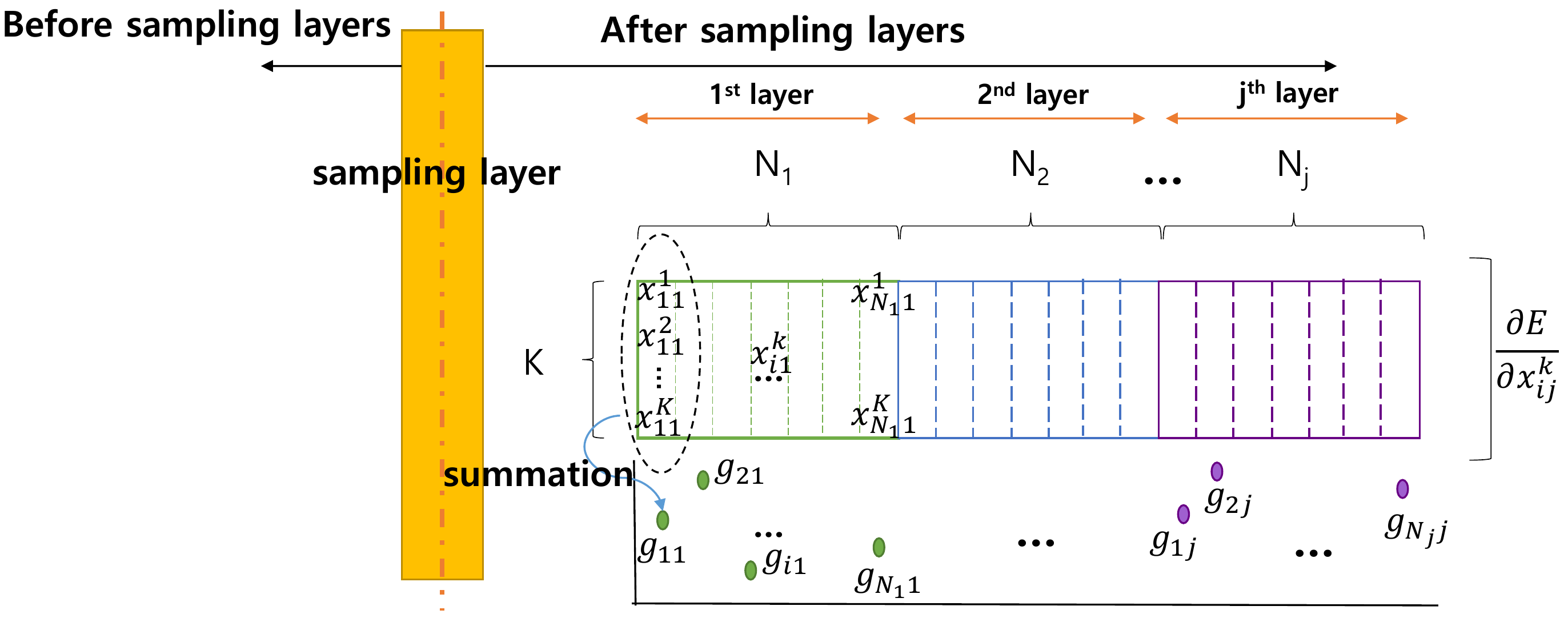}
\end{adjustbox}
\caption{Method of making gradient plot from the gradients $g_i$'s}
\label{fig:grad}

\end{figure}
\setlength{\textfloatsep}{5pt}

For training the proposed segmentation network, considering the large solution space of the network parameters, relatively a small number of samples is provided. Consequently, it is difficult to extract a proper gradient for each layer, and adjusting the layer-wise learning rate is widely used solution for mitigating the effect of the noisy gradient \cite{lecun2012efficient}. In this section, we introduce SPC method for efficient selection of the layer-wise learning rate. 

If we look at CNN as a manufacturing system, we produce a gradient through a process called backpropagation, where the quality of production depends on the parameters controlling the process. If the scattering of the gradients is too large, the learning will not work properly. On the other hand, if the scattering is small, the learning will be done properly and the good performance will be obtained. SPC \cite{shewhart1931economic} is the most popular method for quality control which is based on statistics. SPC monitors data in the current state to understand whether current state is good for producing high quality products. Here, we use the control chart method where the \textit{upper control limit} (UCL) and the \textit{lower control limit} (LCL) are used in the quality control. Both UCL and LCL are set by the mean and the standard deviation of the process and they determine if there is a problem with the process.

In our learning, control chart method is applied to the gradient after 12 epochs. We only apply UCL because we use the absolute value of a gradient. We make a control chart for the same input at the same iteration using each network structure. After sampling, the depth of feature map in the $j$-th layer has $N_j$ dimension. We make a gradient data point $g_{ij}$ for the $i$-th feature map or slice ($i \in \{1,\cdots, N_j\}$) by summing the absolute gradient values of each feature in the slice. This is shown in (\ref{eq:2}) and Figure \ref{fig:grad}. Here, $K$ is the number of features in a slice and $x_{ij}^k$ is the $k$-th feature of the $i$-th feature map in the $j$-th layer.
First, the mean $\mu_j$ and the standard deviation $\sigma_j$ of the gradient are calculated as in (\ref{eq:1}).
Second, $\sigma_j^{\text{low}}$ is the $j$-th layer's standard deviation of the gradients from the experiment of \textit{all low $\gamma$}. UCL is defined using $\mu_j$ and $\sigma_j^{\text{low}}$, as in (\ref{eq:2}). 
The constant $C$ is a parameter for controlling the regularity of the process, and in our method, it is set to 6. 

\begin{equation}
\small
g_{ij} = \sum_{k=1}^{K} || \frac{ \partial E}{\partial x_{ij}^k} ||,  \quad \mu_j = \frac{1}{N_j} \sum_{i=1}^{N_j} {g_{ij}},
\quad \sigma_j = {\sqrt {\frac{1}{N_j} \sum_{i=1}^{N_j} (g_{ij} - \mu_j ) ^2 }},
\label{eq:1}
\vspace{-1.5mm}
\end{equation}

\begin{equation}
\small
\label{eq:2}
UCL = \mu_j + C\sigma_j^{\text{low}} 
\vspace{-1.0mm}
\end{equation}

\begin{figure}[t]
\begin{center}$
\begin{array}{ccc}

\includegraphics[width=4cm, height=2.2cm]{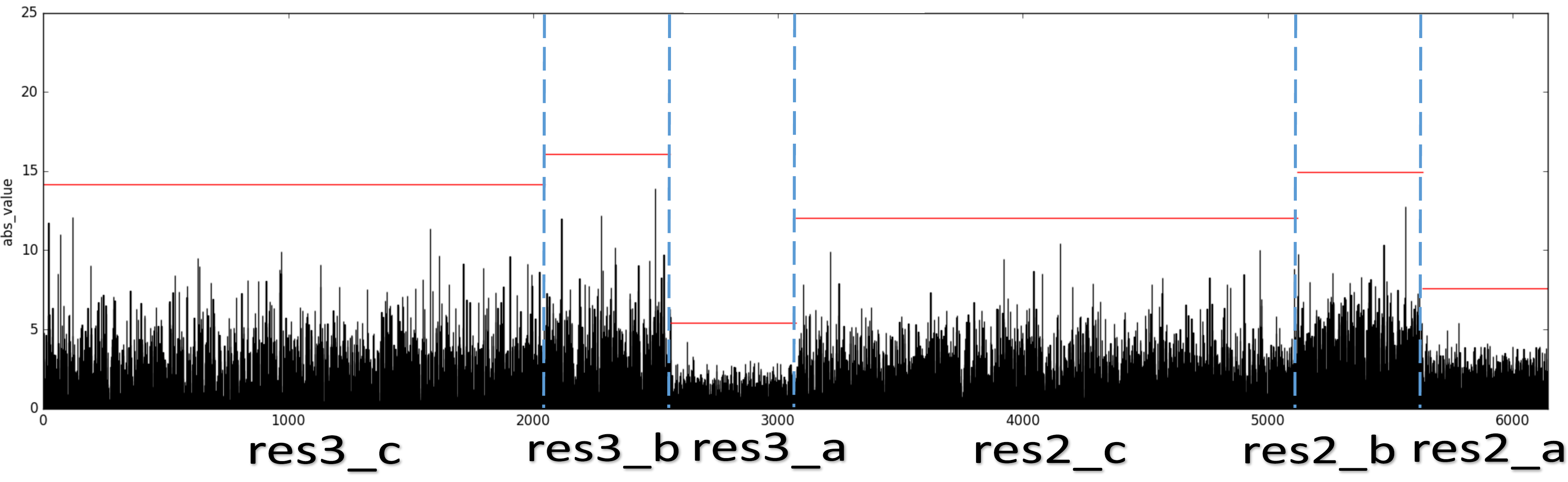}& 	
\includegraphics[width=2cm ,height=2.2cm]{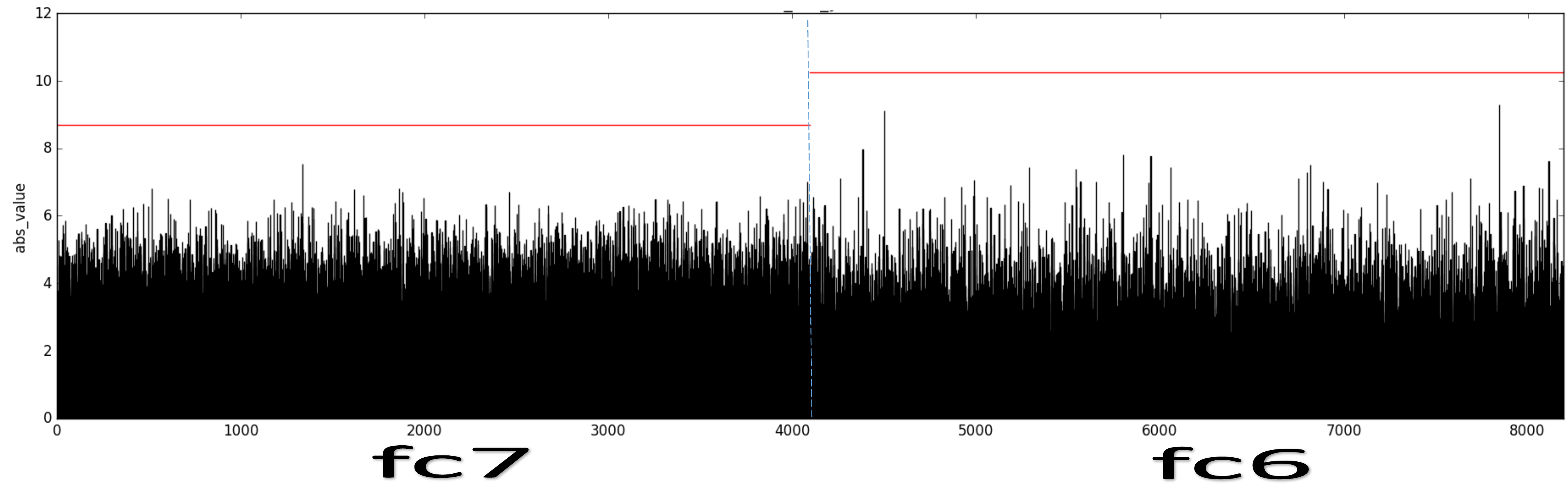}&
\includegraphics[width=4cm]{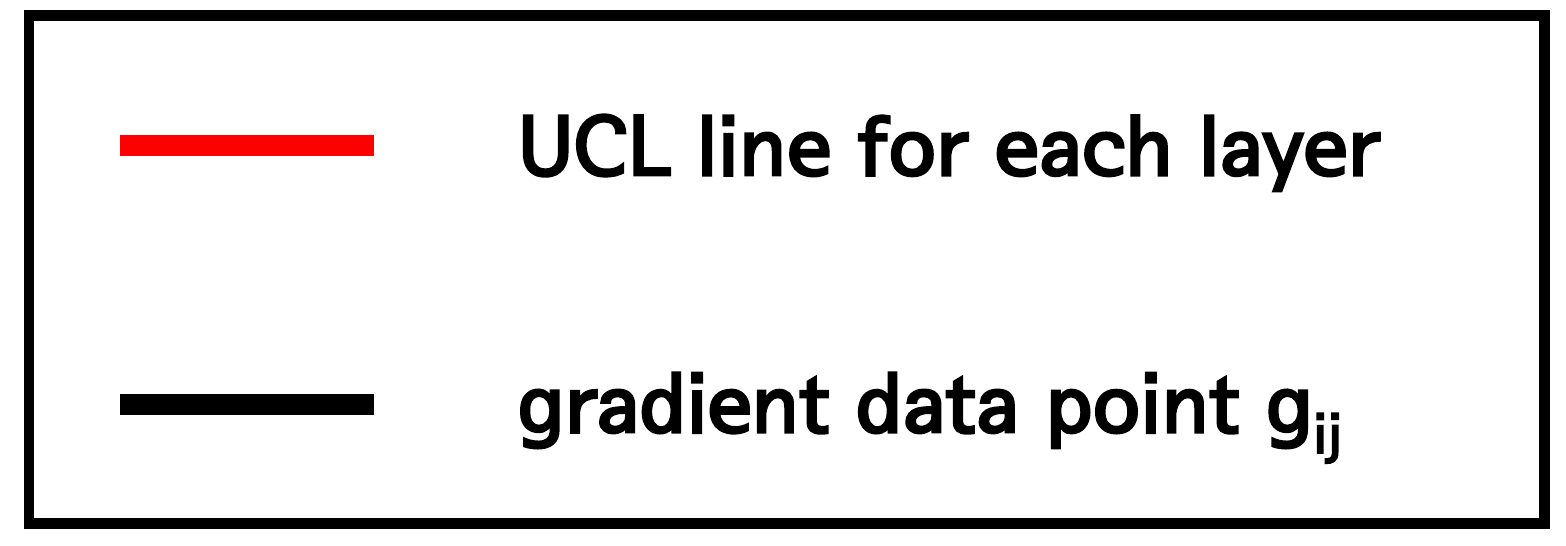}\\ 	
(a)&(b)

\end{array}$
\end{center}

\begin{center}$
\begin{array}{ccc}

\includegraphics[width=4cm,height=2.2cm]{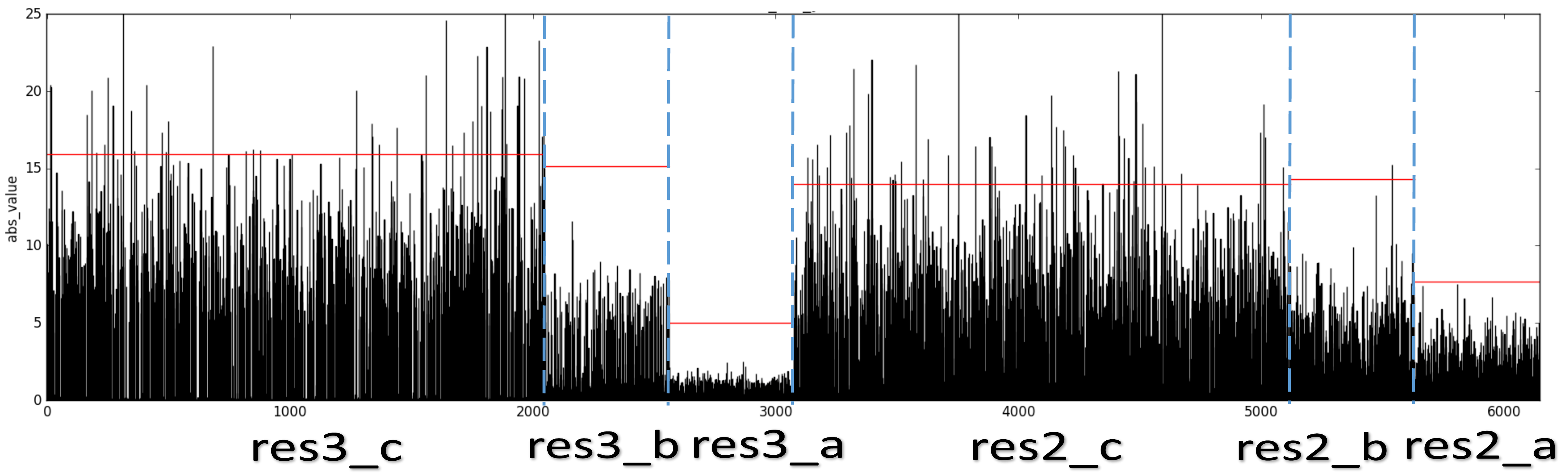}&
\includegraphics[width=2cm,height=2.2cm]{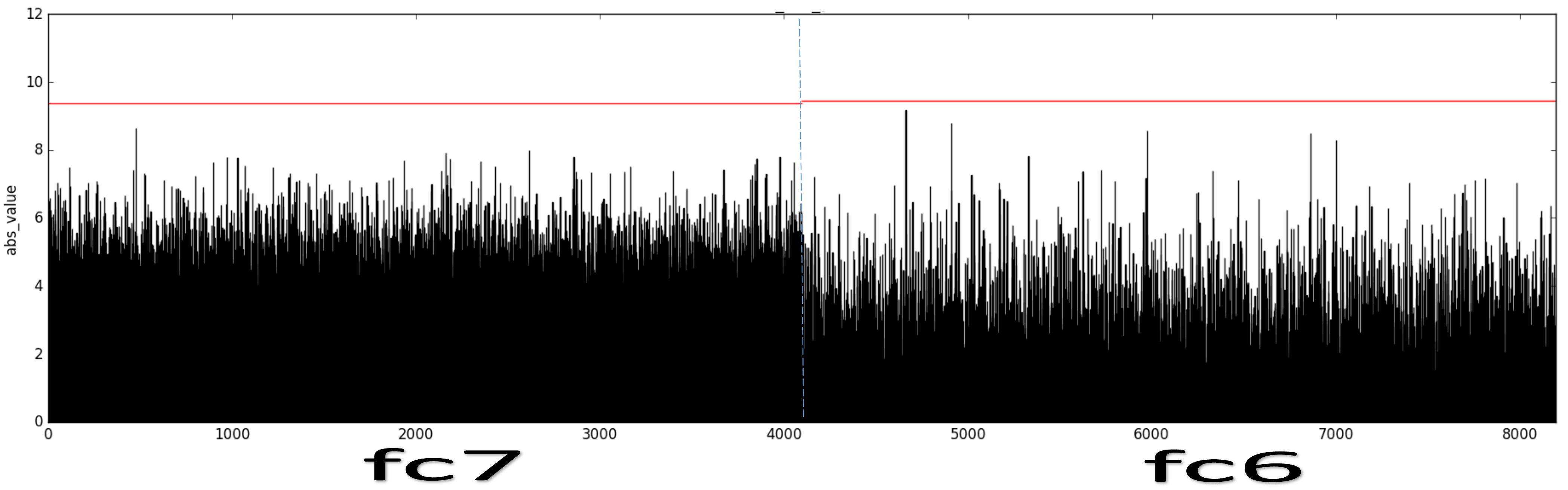}&
\includegraphics[width=4cm,height=2.2cm]{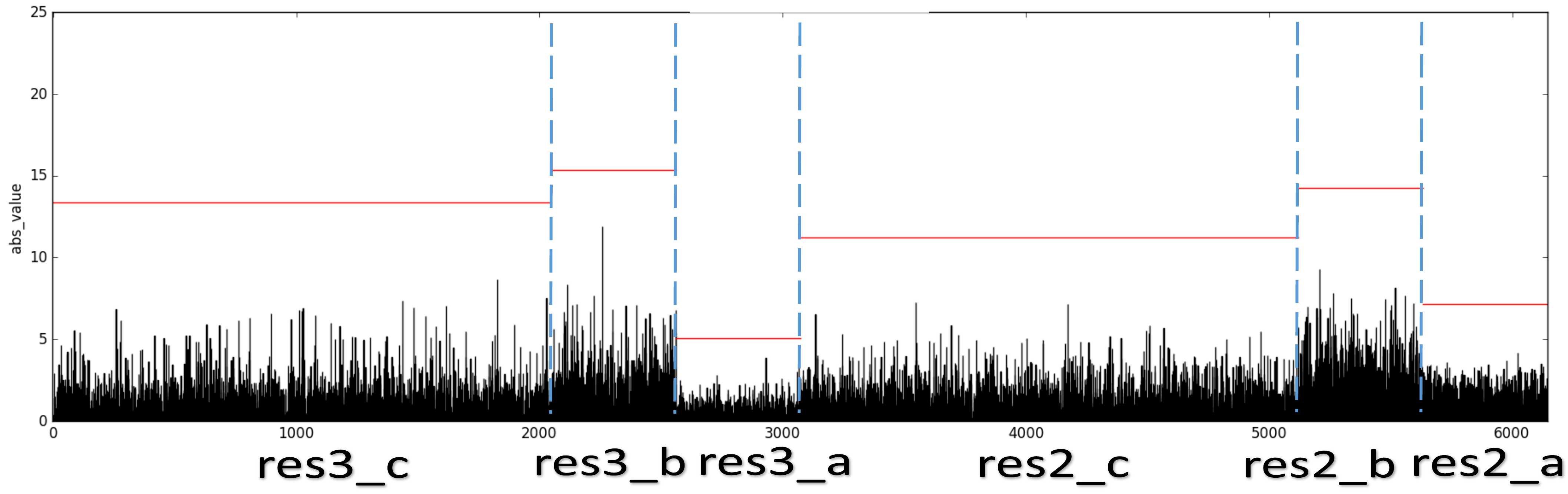}\\
(c)&(d)&(e)

\end{array}$
\end{center}

\vspace{-2mm}
\caption{Gradient plot with UCL. The closer to the origin on the x-axis, the farther away from the input.
(a) all low $\gamma$ gradient plot in Resblock network, (b) all low $\gamma$ gradient plot in FC network, (c) all high $\gamma$ gradient plot in Resblock network, (d) all high $\gamma$ gradient plot in FC network, (e) hybrid $\gamma$ gradient plot in Resblock network }
\label{fig:4}

\end{figure}



Figure \ref{fig:4} shows the plots of the gradient sums $g_{ij}$'s in different layers after the sampling, and the red line in each subfigure represents the UCL line from \textit{all low $\gamma$} standard deviation $\sigma_j^{\text{low}}$.
In Figure \ref{fig:4}(c)(d), we show the gradient plot for the \textit{all high $\gamma$} experiment where the $\gamma$ is set to 10 times the case for the \textit{all low $\gamma$} experiment in Figure \ref{fig:4}(a)(b).
As in figure\ref{fig:4}(c), since the magnitude of $g_{ij}$ exceeds the red line and also starts to fluctuates after passing res3\_c layer, we reduce the learning rate $\gamma$ of the convolution layers placed after res3\_c layer to eliminate noisy gradient from the high learning rate. We note that the learning rate regarding res2\_c is not controlled despite that the gradient of res2\_c is unstable either. It is because the backpropagation for res3\_c is performed before the calculation of gradient for res2\_c. It means that the control of the gradient for res3\_c also brings change of the gradient in res2\_c.  Finally reducing the gradients for the higher layer (res3\_c) automatically brings stable gradients of all the layers. Figure \ref{fig:4}(e) depicts the experiment with hybrid $\gamma$, which clearly shows $g_{ij}$ becomes stable and results in improvement in performance. 
For the FC case, Figure \ref{fig:4}(d) shows less fluctuations in spite of the high value of $\gamma$. In such cases, we make no changes in the learning rate $\gamma$.
\section{Experiment}
\label{sec:exp}
Pascal Context dataset \cite{mottaghi2014role}, which is based on VOC2010 dataset, has  additional detailed categories. 
The most frequently used 59 categories are selected from more than 400 categories and the rest are merged into one category. The number of training and validation images are 4,998 and 5,105, respectively.


SUN-RGBD dataset  \cite{song2015sun} is more difficult to segment than the Pascal Context dataset due to variations in the shape, size and pose of the objects in the same category. 
The dataset contains RGB and depth images from NYU depth v2 \cite{silberman2012indoor}, Berkeley B3DO \cite{janoch2013category}, and SUN3D \cite{xiao2013sun3d} datasets and has 37 indoor environment categories. 
The training set contains 5,285 images, and the test set has 5,050 images.

Here, pixel accuracy (pixel Acc.) measures the ratio of correctly estimated pixels. Mean accuracy (mean Acc.) is the average of category-specific pixel accuracy and mean intersection of union (mean IU) is the average of category-specific ratio of the intersection versus union between the ground truth and the estimation result.

\subsection{Implementation Detail}
\label{sec:exp_sub1}

\begin{table}[tb]
  \centering
  \small
  
    \begin{adjustbox}{width=0.8\textwidth}
    \begin{tabular}{|cccccc||cc|}
    \hline
    \multicolumn{1}{|c|}{1x1,1024} & \multicolumn{1}{c|}{\multirow{4}[4]{*}{Eltwise sum}} & \multicolumn{1}{c|}{1x1,512} & \multicolumn{1}{c|}{\multirow{4}[4]{*}{Eltwise sum}} & \multicolumn{1}{c|}{1x1,1024} & \multirow{4}[4]{*}{Eltwise sum} & \multicolumn{1}{c|}{\multirow{4}[4]{*}{1x1, 4096}} & \multirow{4}[4]{*}{1x1, 4096} \\
    \multicolumn{1}{|c|}{1x1,1024} & \multicolumn{1}{c|}{} & \multicolumn{1}{c|}{1x1,512} & \multicolumn{1}{c|}{} & \multicolumn{1}{c|}{1x1,1024} &       & \multicolumn{1}{c|}{} &  \\
    \multicolumn{1}{|c|}{1x1,2048} & \multicolumn{1}{c|}{} & \multicolumn{1}{c|}{1x1,2048} & \multicolumn{1}{c|}{} & \multicolumn{1}{c|}{1x1,2048} &       & \multicolumn{1}{c|}{} &  \\
\cline{1-1}\cline{3-3}\cline{5-5}    \multicolumn{1}{|c|}{1x1,2048} & \multicolumn{1}{c|}{} & \multicolumn{1}{c|}{} & \multicolumn{1}{c|}{} & \multicolumn{1}{r|}{} &       & \multicolumn{1}{c|}{} &  \\
   \hline
    \multicolumn{2}{|c|}{Block 1} & \multicolumn{2}{c|}{Block 2} & \multicolumn{2}{c||}{Block 3} & \multicolumn{1}{c|}{FC6} & \multicolumn{1}{c|}{FC7} \\
  \hline
    \multicolumn{6}{|c||}{ResBlock}                & \multicolumn{2}{c|}{Fully Convolution} \\
   \hline
    \end{tabular}%
    \end{adjustbox}  
    
  \caption{Network design after sampling layer for semantic segmentation}
  \label{table1}

\end{table}%

All the experiments were performed using the caffe library \cite{jia2014caffe} and we use Pixelnet \cite{bansal2017pixelnet} pre-trained caffe model. For the new layer, we used "xavier" \cite{glorot2010understanding} initialization and dropout \cite{srivastava2014dropout} from the pyramid module. We set the momentum as 0.9 and the weight decay 0.0005. The learning rate started at $10^{-6}$ and decreased to $10^{-7}$ after 16 epoch. The superpixel was created using SLIC\cite{achanta2010slic}.

\subsection{Result from SPC}
\label{sec:exp_sub2}

To analyze the effect of the number of the superpixels between the performance, a cross experiment was conducted as in Table \ref{table2}.
For `Train-$250$s', `Train-$750$s' and `Train-$1600$s', $250$, $750$ and $1600$ superpixels were used.
In the same way, `Test-$250$s', `Test-$750$s' and `Test-$1600$s' were conducted by $250$, $750$ and $1600$ number of superpixels. We used a same network to both $224\times 224$ and $448\times 448$ image size, but only detailed setting of pyramid module was slightly changed. As shown in Table \ref{table2}, the performance was proportional to the number of samples in the train. Surprisingly, the number of superpixels in the test was not significantly affected. That is, as mentioned in Section \ref{sec:prop3}, a small number of sample was provided compared with a large solution space.

 Table \ref{table3} shows performance improvement of tuning learning rate of each layer $\gamma$ with SPC analysis.
In the experiment, \textit{high $\gamma$} was set to $10$ times larger learning rate than \textit{low $\gamma$} case. We note that setting high value of $\gamma$ did not always occur good results. 
By using the proposed statistical process control (SPC) technique, we determined which layer's $\gamma$ should be small. 
The performance was improved when reducing the $\gamma$ value of the last convolution layer before softmax layer in Resblock network. 
In \textit{hybrid $\gamma1$}, the $\gamma$ in last convolution layer was set to $5$ times larger than learning rate than \textit{low $\gamma$} case. The hybrid $\gamma2$ used the equivalent learning rate to \textit{low $\gamma$} case in last convolution layer. 
Consequently, this experiment shows that the higher performance can be achieved by setting learning rate of the last layer $\gamma$ to be smaller than usual case.

\begin{table}[t]
  \centering    
  \begin{adjustbox}{width=\textwidth}
   \begin{tabular}{|c|c|c|c|c||c|c|c|c|c|}
    \hline
    \multirow{2}[4]{*}{image size:$448\times 448$} & \multicolumn{2}{c|}{Train-750s} & \multicolumn{2}{c||}{Train-1600s} & \multirow{2}[4]{*}{image size:$224\times 224$} & \multicolumn{2}{c|}{Train-225s} & \multicolumn{2}{c|}{Train-750s} \\
\cline{2-5}\cline{7-10}          & mean Acc.   & mean IU   & mean Acc.   & mean IU   &       & mean Acc.   & mean IU   & mean Acc.   & mean IU \\
    \hline
    Test-750s & 46.975 & 34.976 & 48.275 & 35.603 & Test-225s & 45.976 & 32.866 & 46.618 & 33.84 \\
    \hline
    Test-1600s & 47.081 & 35.061 & 48.397 & 35.713 & Test-750s & 46.343 & 33.161 & 47.256 & 34.431 \\
    \hline
    \end{tabular}%
    \end{adjustbox}
   
    \caption{Cross experiment between number of sampling and performance }
    \label{table2}
\end{table}%
\setlength{\textfloatsep}{2pt}
  
\begin{table}[t]
  \centering
    \small
     \begin{adjustbox}{width=0.5\textwidth}
    \begin{tabular}{|c|c|c|c|c|}
    \hline
          & all low $\gamma$ & all high $\gamma$ & hybrid $\gamma1$ & hybrid $\gamma2$ \\
    \hline
    mean Acc.   & 50.808 & 50.691 & 51.503 & 51.928 \\
   \hline
    mean IU   & 37.936 & 37.828 & 38.911 & 39.659 \\
   \hline
    \end{tabular}%
    \end{adjustbox}
    
  \caption{Performance of different $\gamma$ policy using Resblock} 
  \label{table3}
   
\end{table}%
\setlength{\textfloatsep}{2pt}

\subsection{Pascal Context}
\label{sec:exp_sub3}
For training and testing the Pascal Context data, we used provided train/validation set to train/test the proposed method.
All the image were resized to $448\times 448$, and $750$ superpixels are used for  each image. 
To analyze the effect of the superpixel method (SLIC), we set the baseline method which uses the same number of the grid dividing the image. 
The `sample(superpixel)' and `sample(grid)' in Table \ref{table4} refers to the case using superpixel and grid for dividing the region of the image.
Both sampling methods generate hypercolumn using conv 3, 4, 5 and 6 in Figure \ref{fig:overview}, but conv 6 was not used for pyramid module.
Instead, pooling with size $14 \times 14$ was used for having same receptive field and convolution filter output feature depth is 4096 for same depth of feature map with HP-SPS.

For segmentation, we used fully convolution layer (FC).
As shown in Table \ref{table4}, mean Acc./mean IU were increased when SLIC is used, and HP-SPS which using pyramid module enhanced the performance more about 2.4\% and 1.5\% , respectively. This means that large receptive field is not the sufficient condition of the high performance. In order to show the effectiveness of SPC, we used two methods each of which uses $3$ Resblocks or $2$ FCs, respectively, mentioned in Table \ref{table1}. 
For Resblock, we do same procedure as described in Section \ref{sec:exp_sub2}. For the \textit{hybrid $\gamma$ FC $10$} case, FC layers will have 10 times higher learning rate after sampling and the performance was stable irrespective of the setting of the $\gamma$. But \textit{hybrid $\gamma$ FC $15$} case, FC layer have 15 times higher than before sampling and we should adjust $\gamma$ from SPC analysis. We reduced the $\gamma$ of `fc7' and it showed slightly better performance.

\begin{figure}[tb]
\centering
\begin{adjustbox}{width =\textwidth}
\begin{tabular}{ccc|ccc}
\bmvaHangBox{\includegraphics[width=2cm, height=3.5cm]{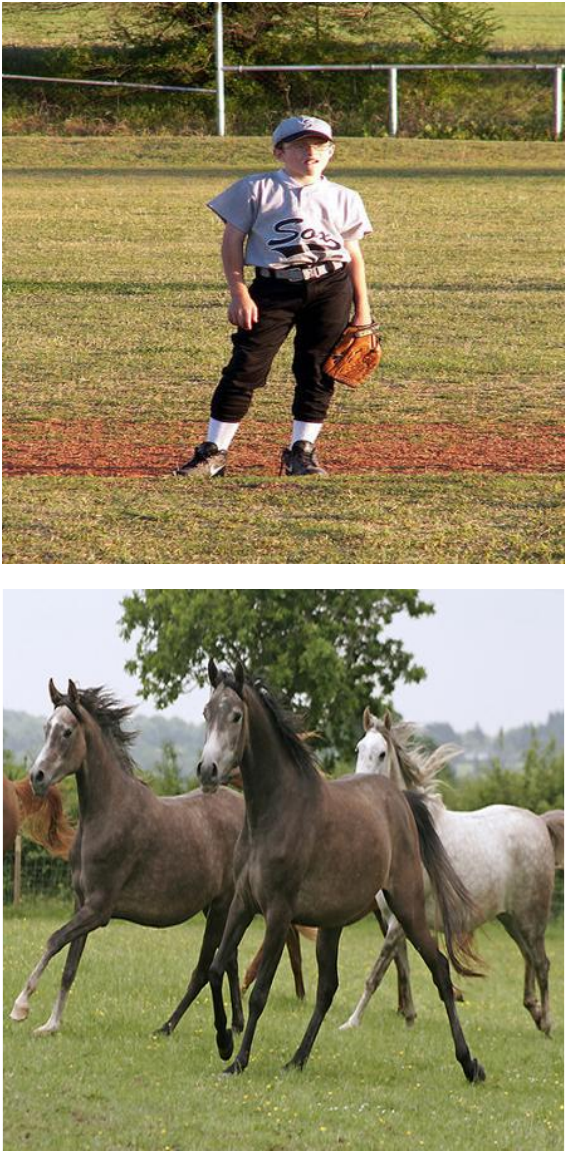}}&
\bmvaHangBox{{\includegraphics[width=2cm, height=3.5cm]{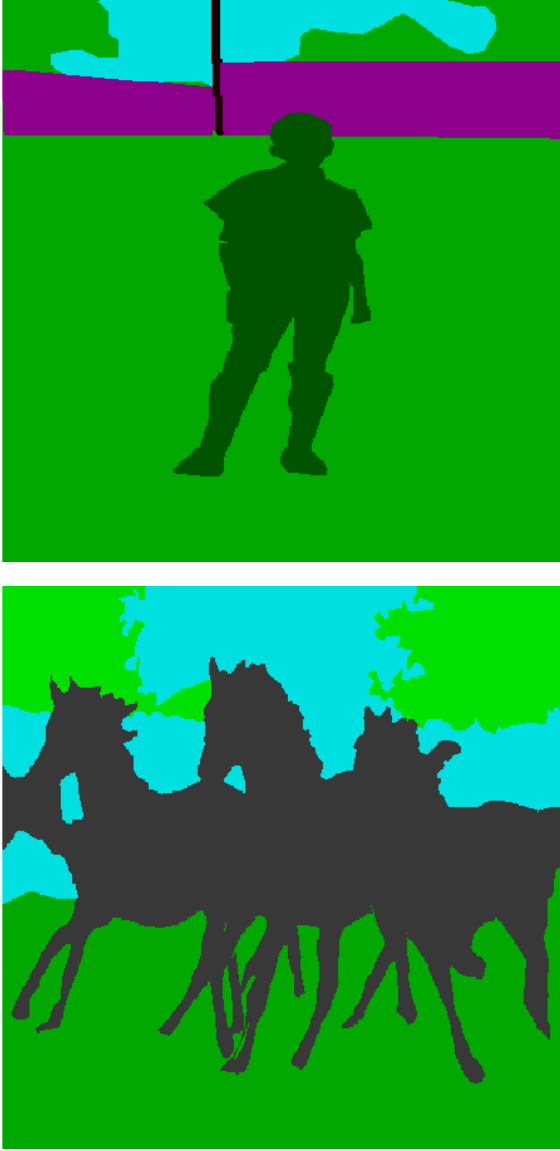}}} &
\bmvaHangBox{\includegraphics[width=2cm, height=3.5cm]{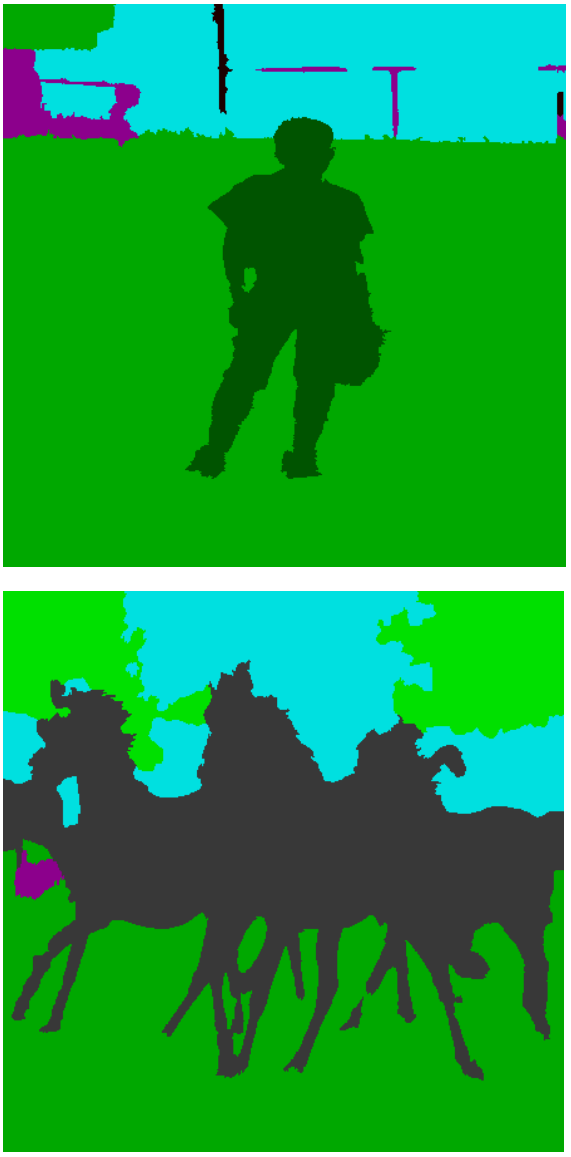}}&
\bmvaHangBox{\includegraphics[width=2cm, height=3.5cm]{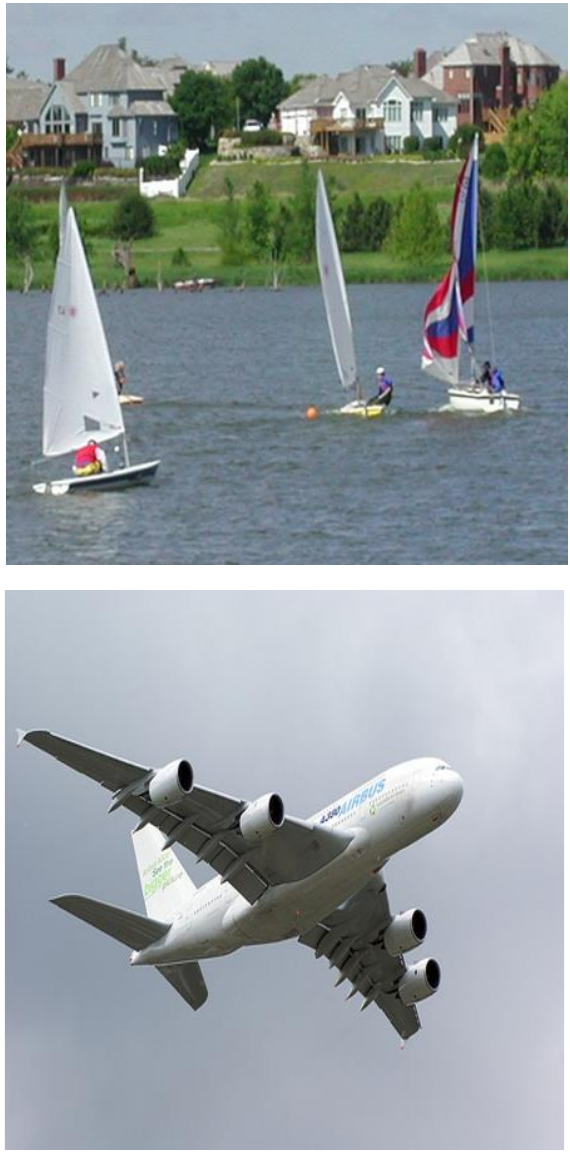}}&
\bmvaHangBox{\includegraphics[width=2cm, height=3.5cm]{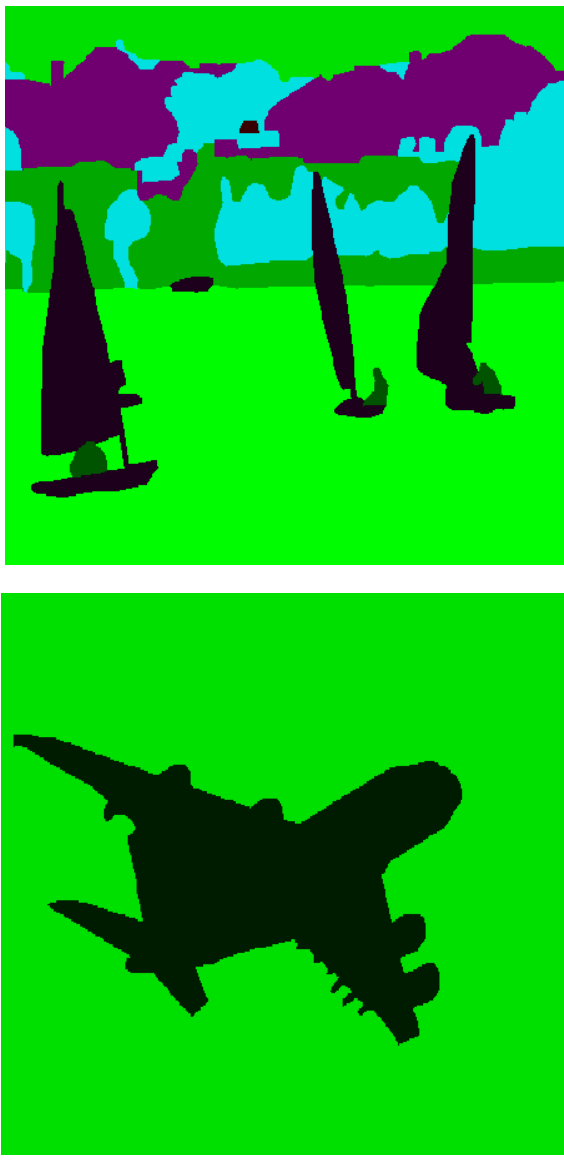}}&
\bmvaHangBox{\includegraphics[width=2cm, height=3.5cm]{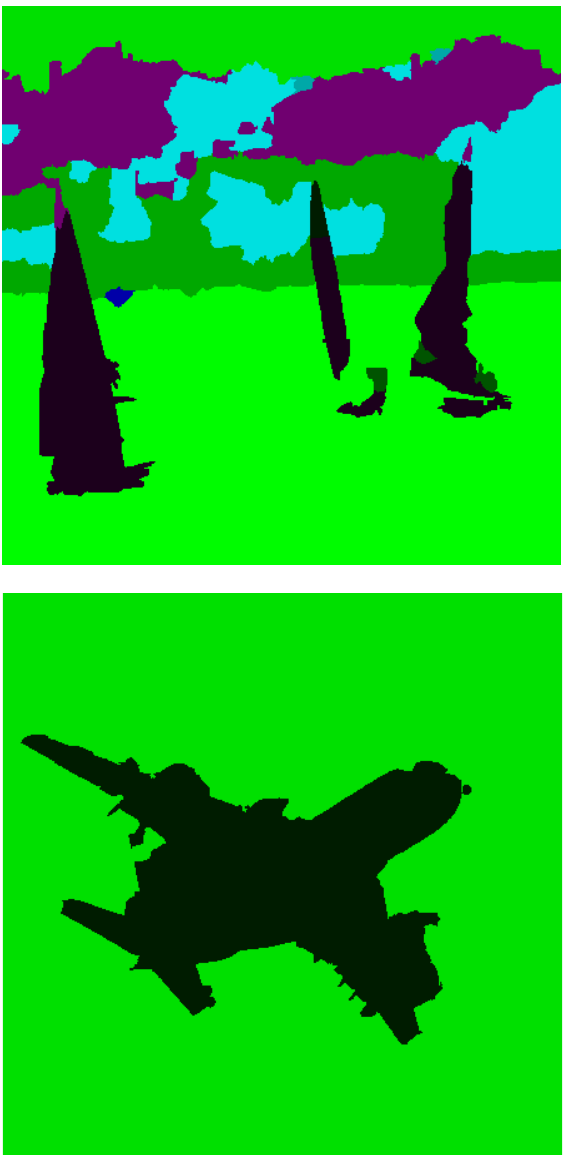}}\\
(a) input& (b) ground truth &(c) our result &(d) input&(e) ground truth&(f) our result
\end{tabular}
\end{adjustbox}
\caption{Some prediction example of our method on Pascal context dataset}
\label{fig:fig6}
\end{figure}

 PixelNet which uses random sampling is related to our train method.
As shown in Table \ref{table4}, Proposed method achieved better mean Acc. but slightly worse mean IU. Compared to the other similar region based method \cite{caesar2016region} applying selective search \cite{uijlings2013selective} to create the region, our method achieved much better performance, as shown in Table \ref{table4}. Figure \ref{fig:fig6} shows several results of our method HP-SPS. 
 
\begin{table}[tb]
  \centering
  \begin{adjustbox}{width =0.95\textwidth}
    \begin{tabular}{|c|c|c||c|c|c|}
    \hline
       proposed method   & mean Acc.   & mean IU   & Others &  mean Acc.      &  mean IU\\
    \hline
    sampling (grid)&  44.735     &  33.179     & FCN-8s\cite{shelhamer2017fully} & 50.7  & 37.8 \\
    \hline
    sampling (superpixel) & 45.536 & 34.457 & DeepLab (without CRF)\cite{chen2016deeplab} &  -     & 37.6 \\
    \hline
    HP-SPS & 47.95 & 35.929 & CRF+RNN \cite{zheng2015conditional} & -      & 39.3 \\
    \hline
    HP-SPS (hybrid $\gamma$, FC 10) & 51.711 & 39.071 & ParseNet \cite{liu15parsenet} &   -    & 40.4 \\
    \hline
    HP-SPS (hybrid $\gamma$, FC 15) & 52.01 & 39.248 & PixelNet\cite{bansal2017pixelnet} & 51.5  & 41.4 \\
    \hline
    HP-SPS(Resblock) & 50.808 & 37.936 & Region based \cite{caesar2016region}& 49.9  & 32.5 \\
    \hline
    HP-SPS (hybrid $\gamma$, Resblock) & 51.928 & 39.659 & IFCN \cite{shuai2016improving} & 57.7  & 45 \\
    \hline
    \end{tabular}%
       \end{adjustbox}
    \caption{Comparison of our results with the baseline methods and others on Pascal context dataset}
 
  \label{table4}%
\end{table}%
\setlength{\textfloatsep}{1pt}

\setlength{\textfloatsep}{1pt}

\subsection{SUN-RGBD dataset}
\label{sec:sun_rgbd}

\begin{figure}[tb]
\centering
\small
\begin{adjustbox}{width=0.9\textwidth}
\begin{tabular}{ccc|ccc}
\bmvaHangBox{\includegraphics[width=2cm, height=6cm]{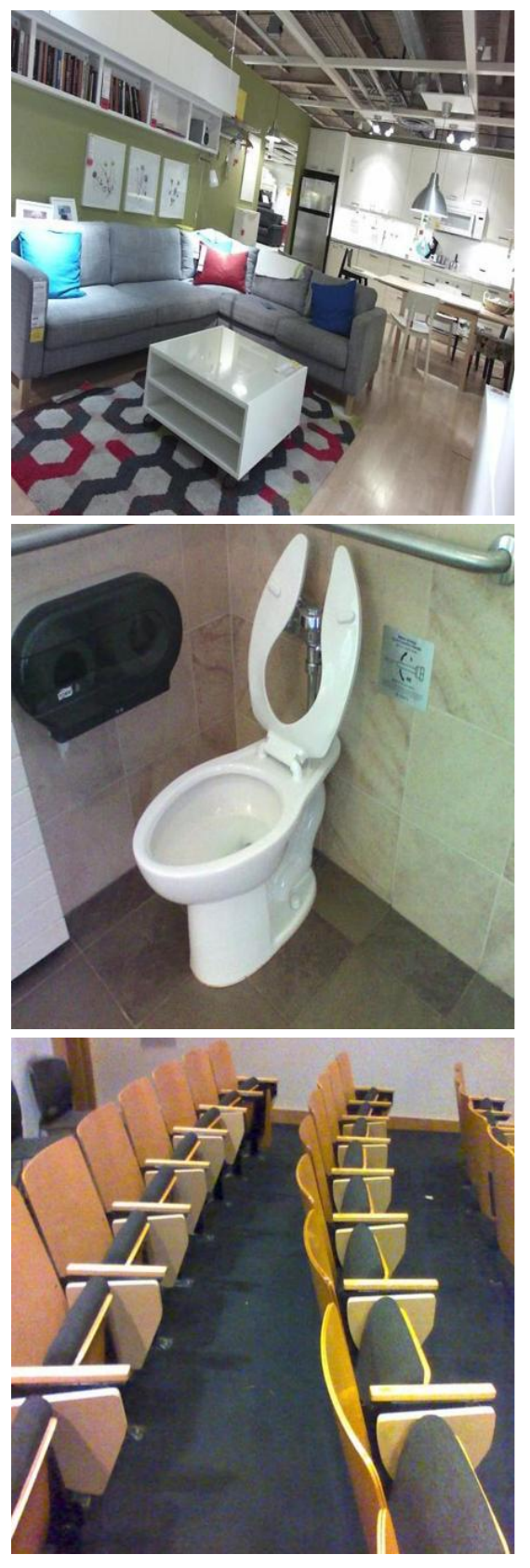}}&
\bmvaHangBox{{\includegraphics[width=2cm, height=6cm]{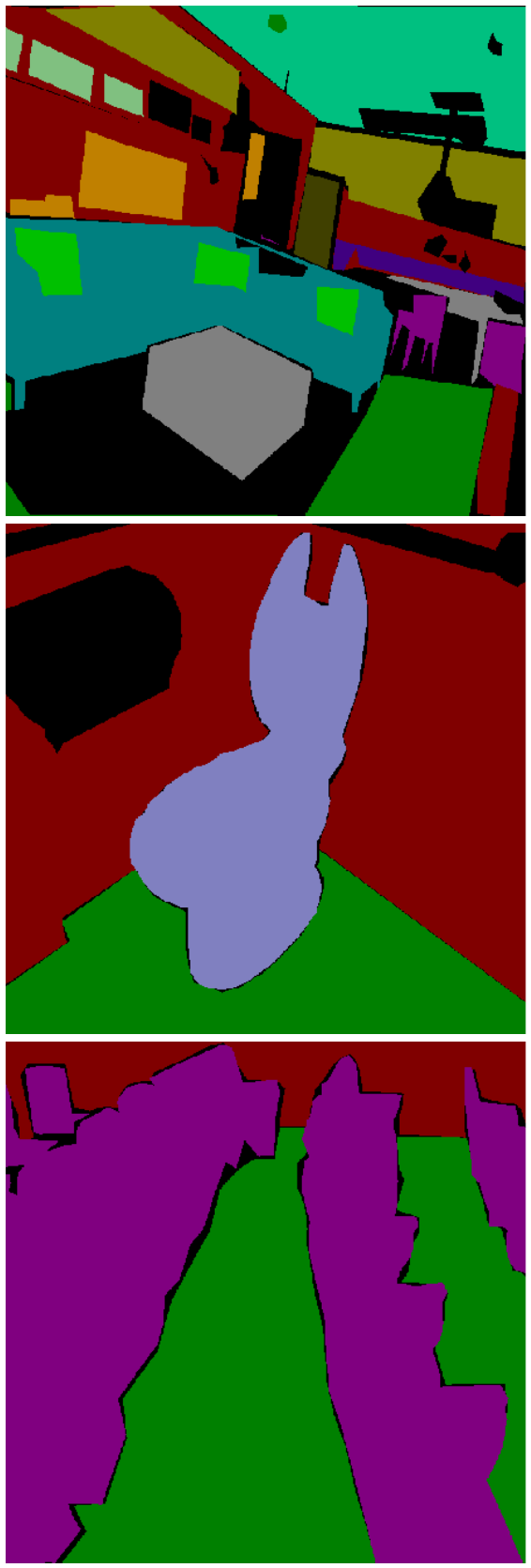}}} &
\bmvaHangBox{\includegraphics[width=2cm, height=6cm]{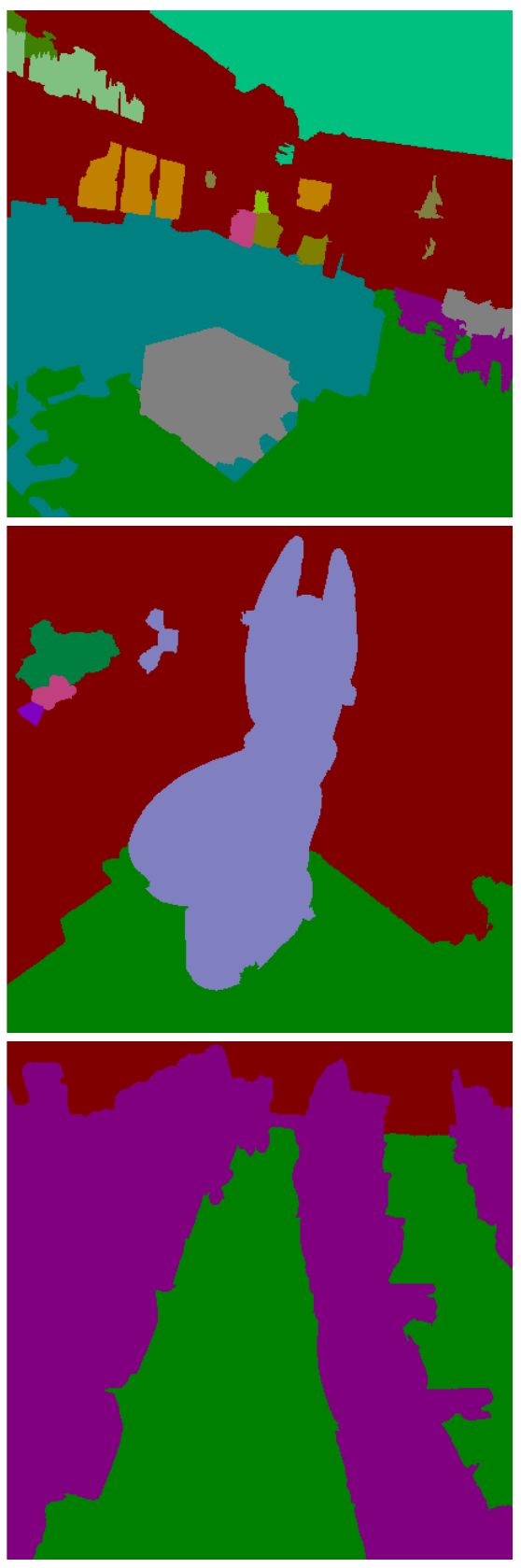}}&
\bmvaHangBox{\includegraphics[width=2cm, height=6cm]{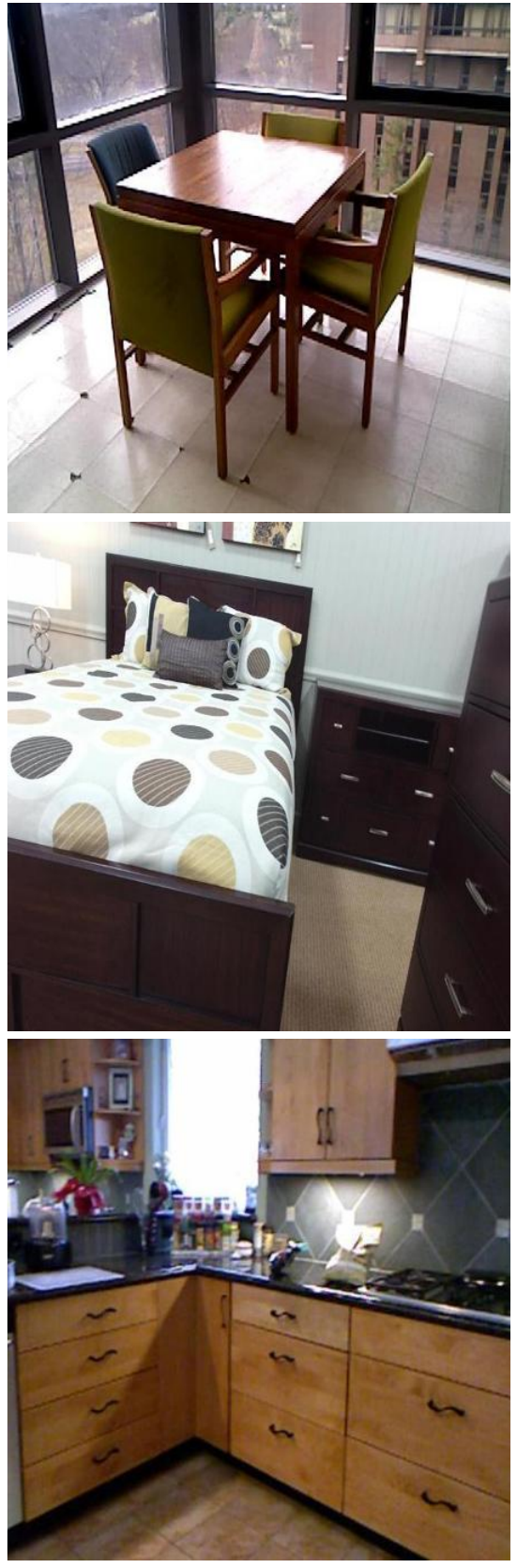}}&
\bmvaHangBox{\includegraphics[width=2cm, height=6cm]{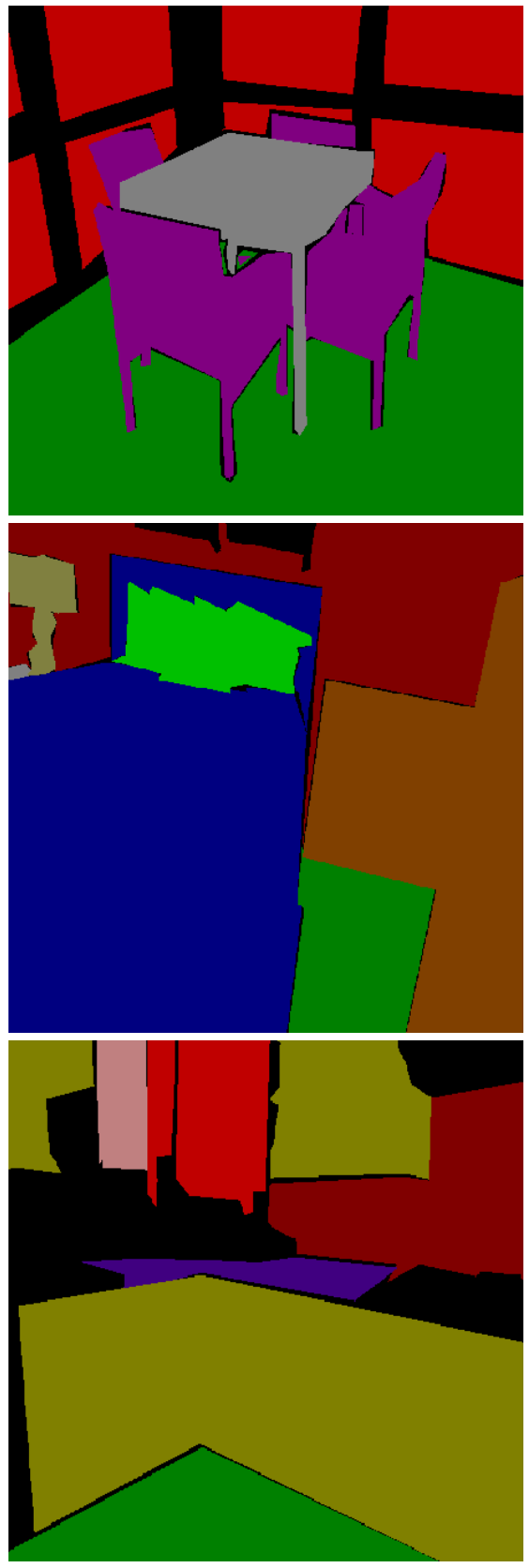}}&
\bmvaHangBox{\includegraphics[width=2cm, height=6cm]{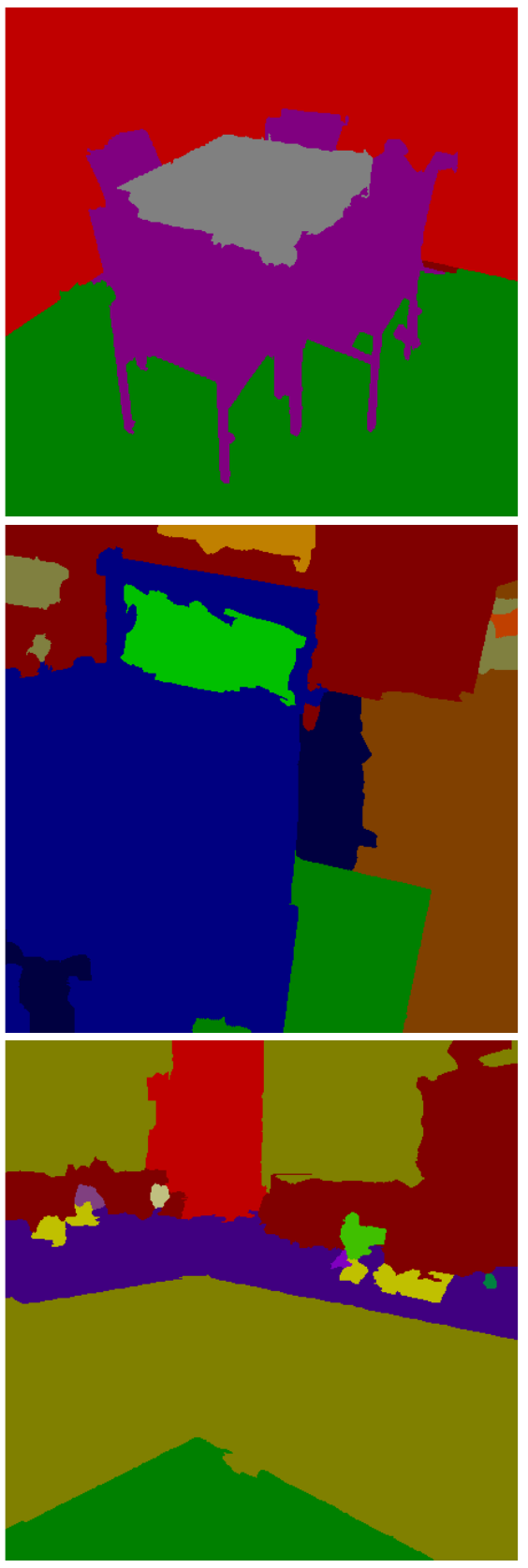}}\\

(a) input& (b) ground truth &(c) our result &(d) input&(e) ground truth&(f) our result

\end{tabular}
\end{adjustbox}
\caption{ Example of our method on SUN-RGBD dataset}
\label{fig:img}
\end{figure}
\setlength{\textfloatsep}{1pt}

Table \ref{tab:sun} shows the performances of various methods. In the experiment, we resized an image to $448\times 448$ and extract 750 samples based on \textit{Simple Linear Iterative Clustering} (SLIC) \cite{achanta2010slic}. We used pre-trained caffe model of pixelNet \cite{bansal2017pixelnet} using only RGB images like other works. Our method used \textit{fully convolution} (FC) layers after sampling for which, a 20-times increased learning rate  was used. 
The method of Lin \etal \cite{lin2016efficient} calculates potential function value by \textit{convolution neural network} (CNN) and applies mean-field approximation for semantic segmentation. In addition, it applies bilinear upsampling to score map and uses \textit{conditional random field} (CRF) to sharpen boundaries. It shows state-of-the-art performance but it uses a very complex model than others. 
IFCN \cite{shuai2016improving} using VGGNet \cite{Simonyan14c} deals with every feature map from pool3 for upsampling.
However, the pixel Acc. of our method and IFCN are not significantly different. The proposed method is better than SegNet \cite{badrinarayanan2015segnet}, FCN \cite{shelhamer2017fully} and DeconvNet \cite{noh2015learning} which applying complex methods for upsampling. 
Also, our method achieved the better performance than DeepLab \cite{chen14semantic} which employing the atrous algorithm for wide receptive field. Figure \ref{fig:img} shows several exemplary results of our method on SUN-RGBD dataset.

\begin{table}[t]
  \centering
 \begin{adjustbox}{width =0.4\textwidth}
    \begin{tabular}{|c|c|c|c|}
    \hline
    \multicolumn{1}{|c|}{method} & \multicolumn{1}{c|}{pixel Acc.} & \multicolumn{1}{c|}{mean Acc.} & \multicolumn{1}{c|}{mean IU} \\
    \hline
    \multicolumn{1}{|c|}{HP-SPS (ours)} & 75.67 & 50.06 & 37.96 \\
    \hline
    Lin \etal \cite{lin2016efficient} & 78.40  & 53.40  & 42.30 \\
   \hline
    IFCN \cite{shuai2016improving}  & 76.90  & 53.46 & 40.74 \\
   \hline
    SegNet \cite{badrinarayanan2015segnet} & 72.63 & 44.76 & 31.84 \\
   \hline
    DeepLab \cite{chen14semantic} & 71.90  & 42.21 & 32.08 \\
   \hline
    FCN \cite{shelhamer2017fully}   & 68.18 & 38.41 & 27.39 \\
   \hline
    DeconvNet \cite{noh2015learning} & 66.13 & 33.28 & 22.57 \\
   \hline
    \end{tabular}%
    \end{adjustbox}
     \caption{SUN-RGBD result. DeepLab, FCN and DeconvNet results are copied from \cite{badrinarayanan2015segnet} }
  \label{tab:sun}
  \vspace{-1.0em}
\end{table}%

\subsection{Speed according to the number of sampling }
\label{sec:time}

We reduce the number of sample 
, and get better speed than pixelNet. 
With using Intel (R) Core (TM) i7-4790D CPU 4.00 GHz without GPU, PixelNet used about $30.33$ seconds on $224\times 224$ image. FCN-8s, One of the popular works, took $6.85$ second and SegNet \cite{badrinarayanan2015segnet} marked $5.4$ second on $448\times 448$ input. We evaluated the performances using various number of superpixes for Pascal Context dataset on $448\time 448$ image. Trained model with 750s superpixels was used for all experiment in Table \ref{tab:time}.

\begin{table}[tb]
  \centering
  \begin{adjustbox}{width=0.4\textwidth}
    \begin{tabular}{|c|c|c|c|}
    \hline
          & \multicolumn{1}{c|}{mean ACC} & \multicolumn{1}{c|}{mean IU} & \multicolumn{1}{c|}{time(sec)} \\
    \hline
    sp(250) & 51.33 & 38.51 & 2.4 \\
    \hline
    sp(750) & 52.01 & 39.25 & 3.1 \\
    \hline
    sp(1600) & 52.16 & 39.42 & 4.4 \\
    \hline
    \end{tabular}%
     \end{adjustbox}
      \caption{ Computational CPU time and performance on $448\times 448$ input}
  \label{tab:time}%
\end{table}%
\setlength{\textfloatsep}{1pt}

\section{Conclusion}
\label{sec:conclusion}

Because most methods in semantic segmentation perform pixel-wise classification, there are many redundancy operations in both train and test. 
More specifically, because neighboring pixels have a high probability to be the same class, unnecessary operations are needed to estimate semantic category on all the pixels.
In addition, there are also unnecessary operations in the feature extraction that requires a smaller feature maps enlarged to the size of the original image. 
Also, in the training phase, they do not meet the IID assumption of SGD because neighboring pixels are highly correlated. 
This paper comprehensively solves these problems by using superpixel-based sampling and uses hypercolumn with pyramid module for robust feature representation. 
Besides, since only 0.374\% of the pixels are sampled, a learning problem arises, which is solved through statistical process control.
We evaluated the proposed method on the Pascal Context and SUN-RGBD dataset and compared the performance with similar methodologies. The proposed method shows equal or better performance and is more efficient than the compared methods.

\section{Acknowledgement}
\label{sec:Ack}
The research was supported by the Green Car development project through the Korean MTIE (10063267) and ICT R\&D program of MSIP/IITP (2017-0-00306).

\end{document}